\documentclass[preprint]{vgtc}               

\graphicspath{{figures/}{pictures/}{images/}{./}} 
\usepackage{amsmath}
\usepackage[noend,linesnumbered,ruled,vlined]{algorithm2e} 
\usepackage{color}
\usepackage{multirow}
\usepackage{tabularray}
\usepackage{amsmath}
\usepackage{amssymb}
\usepackage{enumitem}

\usepackage{times}                     
 
\usepackage{tabu}                      
\usepackage{booktabs}                  
\usepackage{lipsum}                    
\usepackage{mwe}                       
\usepackage{comment}                     

\usepackage{mathptmx}                  

\onlineid{1114}

\vgtccategory{Research}

\vgtcinsertpkg

\title{FovealNet: Advancing AI-Driven Gaze Tracking Solutions for Optimized Foveated Rendering System Performance in Virtual Reality}

\author{Wenxuan Liu, Monde Duinkharjav, Qi Sun, Sai Qian Zhang\thanks{Corresponding author}}
\affiliation{\scriptsize Tandon School of Engineering, New York University}

\teaser{
  \centering
  \includegraphics[width=\linewidth]{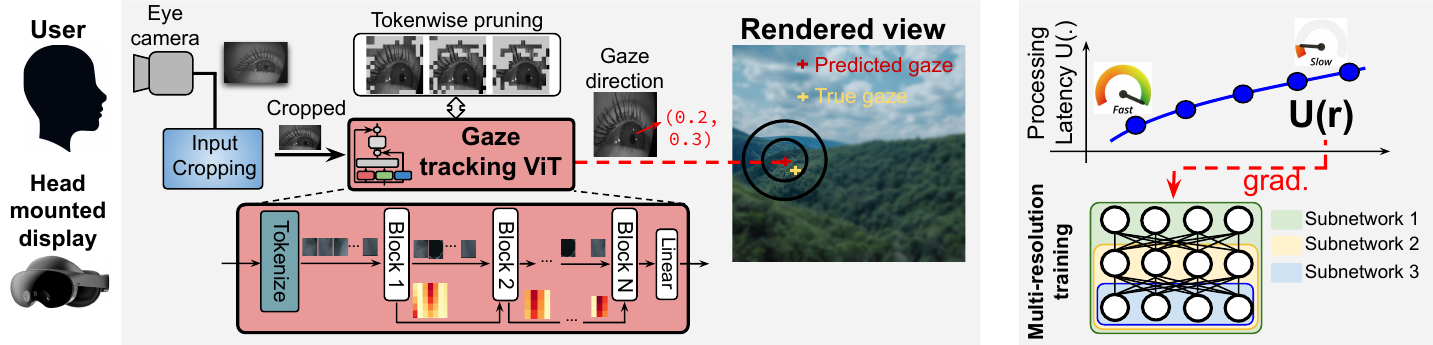}
  \caption{FovealNet for efficient gaze-tracked foveated rendering system operation.}
  \label{fig:intro_figure}
}

\abstract{
    Leveraging real-time eye-tracking, foveated rendering optimizes hardware efficiency and enhances visual quality virtual reality (VR). This approach leverages eye-tracking techniques to determine where the user is looking, allowing the system to render high-resolution graphics only in the foveal region—the small area of the retina where visual acuity is highest, while the peripheral view is rendered at lower resolution. However, modern deep learning-based gaze-tracking solutions often exhibit a long-tail distribution of tracking errors, which can degrade user experience and reduce the benefits of foveated rendering by causing misalignment and decreased visual quality.

    This paper introduces \textit{FovealNet}, an advanced AI-driven gaze tracking framework designed to optimize system performance by strategically enhancing gaze tracking accuracy. To further reduce the implementation cost of the gaze tracking algorithm, FovealNet employs an event-based cropping method that eliminates over $64.8\%$ of irrelevant pixels from the input image. Additionally, it incorporates a simple yet effective token-pruning strategy that dynamically removes tokens on the fly without compromising tracking accuracy. Finally, to support different runtime rendering configurations, we propose a system performance-aware multi-resolution training strategy, allowing the gaze tracking DNN to adapt and optimize overall system performance more effectively.
    Evaluation results demonstrate that FovealNet achieves at least $1.42\times$ speed up compared to previous methods and 13\% increase in perceptual quality for foveated output.
} 


\begin{document}


\firstsection{Introduction}

\maketitle

Human visual acuity varies across the visual field. The fovea, the central region of the retina, is responsible for our sharpest vision. This region, although small, is densely packed with photoreceptor cells, allowing us to perceive fine details and vibrant colors within our direct line of sight. As we move away from the fovea, our visual acuity decreases rapidly, meaning that the peripheral vision is less sensitive to fine details. 
Foveated rendering leverages this phenomenon by allocating more computational resources to the fovea while reducing detail in the periphery. This technique significantly enhances system performance by lowering the rendering workload without compromising the perceived visual quality, making it a critical innovation for applications such as virtual reality (VR)~\cite{patney2016perceptually,albert2017latency,franke2021time}, video encoding~\cite{kaplanyan2019deepfovea,illahi2020interplay,illahi2021foveated} and gaming~\cite{hegazy2019content,illahi2020cloud,zou2021enhancing}.
By aligning rendering fidelity with human gaze patterns, foveated rendering optimizes both visual experience and computational efficiency.

Therefore, VR systems commonly require gaze-tracking for foveated rendering, which is usually fulfilled with a deep neural networks (DNNs). By precisely determining the user's point of focus in real-time, the gaze-tracked foveated rendering (TFR) system can precisely catch the location of the foveal region which is rendered with the highest resolution, followed by the \textit{inter-foveal region} and \textit{peripheral region}, which then will be rendered from fine to coarse level, as shown in the left part of \cref{fig:intro_figure}.

Accurate gaze tracking is fundamental to the successful implementation of TFR. Without reliable gaze tracking, the system cannot accurately adapt to the user's visual focus, leading to potential misalignments between rendered detail and real gaze position, which can result in noticeable artifacts and diminished user experience. Therefore, integrating robust gaze tracking mechanisms is imperative for optimizing performance and ensuring seamless, high-fidelity visuals in TFR. Although several prior studies have proposed AI-based gaze tracking solutions that perform well on test datasets~\cite{deepvog, vrpaper_zhu, etracker, angelo2021vrpaper, Kothari2020EllSegAE}, our experiments in this paper show that these solutions can considerably reduce the efficiency of TFR. This is because, despite having low average gaze tracking errors, the errors often follow a long-tail distribution, resulting in substantial inaccuracies in detecting the user's gaze location in various scenarios. These errors can further cause the rendered foveal region to be misaligned with the user’s actual gaze, leading to decreased visual quality and undermining the intended performance gains of foveated rendering, ultimately diminishing the user experience.


To tackle this challenge, we introduce a novel training approach that integrates TFR system performance directly into the gaze tracking DNN's training process, thereby optimizing the overall performance. Specifically, we focus on minimizing system latency in this work, as latency is a key factor in VR environments. Moreover, our approach can be extended to optimize various system performance metrics in different TFR scenarios (e.g., power consumption).

Furthermore, previous studies have highlighted the significance of implementation overhead for gaze-tracking DNNs~\cite{vrpaper_zhu, albert2017latency}, as this additional cost can often outweigh the performance benefits gained from TFR. To reduce the computational complexity of gaze tracking DNN, we develop a simple approach that focuses on efficiently capturing the eye region centered around the pupil, minimizing computation for irrelevant peripheral pixels. This event-driven design also enables efficient reuse of buffered gaze-tracking results during the execution. In addition, we introduce fine-grained pruning mechanisms targeting input tokens within the gaze-tracking model, reducing unnecessary computation in non-informative areas such as the eyelashes. 
\begin{figure}[t]
\centering
\includegraphics[width=1\linewidth]{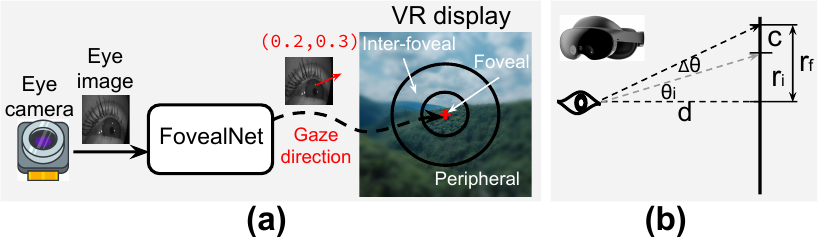}
\caption{(a) TFR system configuration. (b) Foveated rendering in VR device.}
\label{fig:gaze_tracking}
\vspace{-8pt}
\end{figure}

Finally, the hardware processing latency for image rendering and tracking often demonstrates dynamic behavior, influenced by user modifications to system settings and resource allocation with other applications. This variability necessitates a dynamic configuration of the gaze tracking DNN to ensure optimal system performance. In address this, we introduce a multi-resolution DNN training framework that trains the gaze-tracking DNN across various configurations simultaneously, as shown in the right part of \cref{fig:intro_figure}. During operation, the most suitable DNN configuration is chosen based on the current system conditions, facilitating optimal dynamic performance for TFR. Our contributions are summarized as follows:

\begin{itemize}[leftmargin=*]
    \item We propose~\textit{FovealNet}, an AI-driven gaze tracking solutions for optimized system performance of TFR. FovealNet employs a performance-aware training strategy by directly optimizing the TFR system latency during its training phase.
    \item To reduce the implementation cost of FovealNet, we introduce an efficient input cropping method that focuses on extracting the central eye region. Furthermore, we propose an adaptive input token pruning technique for the transformer-based gaze-tracking DNN, achieving superior computational efficiency while maintaining the tracking performance.
    \item To accommodate variations in TFR system configurations, we propose a multi-resolution DNN training framework that allows the resulting gaze-tracking DNN to dynamically adapt by selecting the optimal configuration based on current system conditions.
\end{itemize}

\section{Background and Related Work}
\label{sec:background}
\subsection{Gaze Tracking Algorithms}
Generally, gaze tracking methods can be broadly classified into model-based and appearance-based approaches~\cite{Overview_tracking, Over_tracking_1}. Model-based techniques estimate gaze direction by utilizing a 3D eye model that mimics physiological structures~\cite{Model_based_method, Model_based_method_1, Model_based_method_2, lu2022model}. These approaches generally involve two stages: (1) employing an eye feature extraction neural network to produce salient eye features and fitting a geometric eye model, and (2) predicting the gaze direction based on the fitted eye model. Essentially, model-based methods transform the gaze tracking problem as an eye segmentation task. Most of model-based approaches utilize U-Net with convolutional operations for eye feature extraction, with most of the computations arising from this process~\cite{vrpaper_zhu, deepvog, etracker,zhang2024swifteye}. Although previous studies have shown high accuracy in the eye segmentation task~\cite{RITnet}, the geometric methods used to derive gaze direction from the fitted eye model can be prone to inaccuracies, inherently introducing an error of more than $2^\circ$, compared to the ground truth. 
This is primarily due to two reasons: (1) inaccuracies in fitting the eye center and radius during the eye model initialization phase, and (2) during the optimization phase, the geometric model imposes additional constraints on the potential shapes and positions of the pupils, further lead to mismatches between the output and the ground truth~\cite{Model_based_method}.

In contrast, appearance-based gaze tracking methods directly use eye images as input and learn a mapping from these images to gaze direction~\cite{Overview_tracking, appearance_zhang_1, appearance_zhang_2}. Compared to model-based approaches, these methods typically require larger amounts of training data. The scale and complexity of the required training data have led to the development of a wide range of learning-based techniques, including linear regression~\cite{linear_regression_2}, random forests~\cite{random_forest_1}, and k-nearest neighbors (KNN)~\cite{random_forest_1, knn_1}, and CNNs~\cite{resnet_inception,mazzeo2021openeds}. Vision Transformers (ViTs)~\cite{dosovitskiy2020image}, renowned for their state-of-the-art performance in various vision tasks, have also been adopted for gaze tracking~\cite{nvgaze}.

However, the significant computational demands of ViTs present a major challenge for real-time gaze tracking. To tackle this, we propose a system performance-driven, multi-resolution gaze tracking transformer network. To the best of our knowledge, this work represents the first attempt to train a gaze tracking DNN while considering the system performance of TFR.

\subsection{Gaze-Tracked Foveated Rendering}
\begin{figure}[tp]
\centering
\includegraphics[scale=0.7]{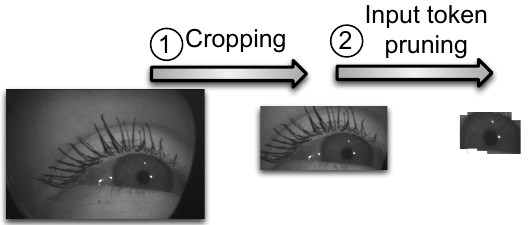}
\caption{Given an input eye image, FovealNet first crops the image to remove background patches (step 1). The remaining patches undergo fine-grained token-wise pruning to eliminate unimportant tokens (step 2), and the remaining tokens are processed by the ViT.}
\label{fig:token-pruning}
\vspace{-10pt}
\end{figure}

The growing popularity of virtual reality has led to a significant demand for rendering high-resolution images on resource-limited hardware platforms, such as head-mounted displays (HMDs)~\cite{albert2017latency, foveated_rendering_basic}. In these contexts, minimizing system latency is crucial to prevent visual distortions and visual artifacts. If the system fails to keep up, users may experience a disconnect between what they see and what they feel, leading to a quality degradation of user experience~\cite{latency_effect2, latency_effect_3}. Early HMDs employed full-resolution rendering, where each pixel within the user’s field of view was rendered uniformly. While this method maintained visual quality, it resulted in unnecessary computational overhead and increased system latency.

TFR is a VR technique that optimizes rendering using gaze-tracking, typically powered by DNNs. It detects the user's gaze location in real-time, rendering the~\textit{foveal region} at the highest resolution (\cref{fig:gaze_tracking}(a)), while surrounding areas, namely the~\textit{inter-foveal region} and~\textit{peripheral region}, are rendered from fine to coarse level without introducing visual artifacts~\cite{foveated_rendering_basic, latency_effect_3, ye2022rectanglefr}. 
As depicted in \cref{fig:gaze_tracking}(b), in TFR, the radius $r_{f}$ of the foveal region (in pixels) is set based on $r_{f} = r_{i} + c = \rho d \cdot \tan(\theta_{i} + \Delta\theta) = d\tan(\theta_{f})$~\footnote{This formula is derived by assuming the gaze is positioned at the center of the front view, representing the maximum radius of the rendering region.}, where $\rho$ represents the display's pixel density, $d$ is the distance between the user and the screen, $\theta_{i}$ is the eccentricity angle subtended by the fovea, and $\theta_{f}=\theta_{i} + \Delta\theta$ is the resultant eccentricity angle that accounts for the gaze tracking error $\Delta\theta$. The radius of the foveal region without tracking error is $r_{i} = d\tan(\theta_{i})$, an error constant $c = d\tan(\theta_{f}) - d\tan(\theta_{i})$ is used to represent the changes on foveal region radius caused by $\Delta\theta$.
The value of $\theta_{i}$ is typically adjusted differently~\cite{foveal_definition,nvidia_vrs,albert2017latency} depending on the experimental results from user studies. For example, in~\cite{foveal_definition}, $\theta_{i}$ is set to 5.2° around the gaze location, while the inter-foveal region covers the area from 5.2° to 17°. In the context of HMD, the distance between the user and the screen $d$ is remain fixed.

\begin{figure}[t]
\centering
\includegraphics[width=1\linewidth]{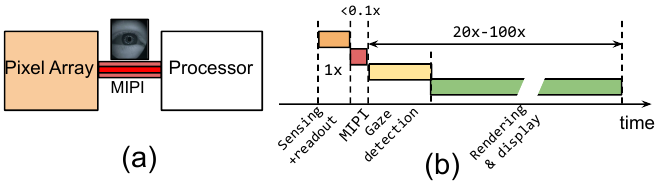}
\caption{(a) TFR system configuration. (b) Latency breakdown (normalized) of TFR.}
\label{fig:tfr_latency}
\end{figure}
\subsection{DNN Pruning}


Pruning techniques are commonly used in DNNs to reduce memory and computational costs. Beyond weight pruning, research has focused on pruning intermediate tokens in ViT~\cite{VIT,karpov2022vit}s. For instance, SPViT~\cite{spvit} removes redundant tokens with a token selector; S$^{2}$ViTE~\cite{s2vite} uses sparse training to prune tokens and attention heads; and Evo-ViT~\cite{evo-vit} employs a slow-fast token evolution mechanism to retain essential information.

Although previous work has utilized cropping techniques to remove redundant surrounding pixels from eye images~\cite{vrpaper_zhu}, the specific characteristics of the gaze-tracking task allow us to further eliminate redundant input tokens within near-eye images, such as those representing the eyelashes. Compared to tokens depicting the iris and pupil, these elements are relatively irrelevant to gaze tracking results~\cite{Model_based_method}. This insight led us to implement fine-grained token-wise pruning on top of the cropped input of ViT and its intermediate activations (step 2 in \cref{fig:token-pruning}). Our proposed token-wise pruning approach ranks input tokens based on their importance scores (attention scores) with respect to the final gaze prediction and eliminates unimportant tokens, leading to a significant reduction in computational costs with negligible accuracy impact.



\section{Motivation}
\label{sec:motivation}
\subsection{TFR System and Pipeline}
\label{sec:latency_breakdown}
A TFR system in modern VR devices (e.g., HMDs) usually comprises three main components (\cref{fig:tfr_latency}(a)): a near-eye camera, a host processor, and an interconnection link (MIPI~\cite{lancheres2019mipi}). A typical TFR pipeline is shown in \cref{fig:tfr_latency}(b). The process begins with capturing an eye image using a near-eye monochrome camera. This image is then preprocessed by the image signal processor (ISP) and readout before being transferred to the host processor via the MIPI connection. After the host processor receives the image, it is sent to the eye-tracking DNN, which estimates the gaze direction. This estimated gaze direction is then utilized to guide the foveated rendering process to produce the rendered VR scene.

\cref{fig:tfr_latency} (b) also provides an approximate latency breakdown of the TFR process. Camera sensing and MIPI communication account for a small fraction of the total latency, approximately 1 ms~\cite{you2023eyecod,liu20204,angelopoulos2020event,sun2024estimating} and less than 1 ms~\cite{lee20196,mipi}, respectively. In contrast, the gaze detection, along with the subsequent rendering and display process, usually consumes a much larger portion (20-100$\times$ longer) of the overall latency, based on the studies from~\cite{albert2017latency, power_quality_tradeoff}. 


\subsection{Foveated Visual Quality Assessment}
\label{sec:fvvdp}
\begin{figure}[t]
\centering
\includegraphics[width=1\linewidth]{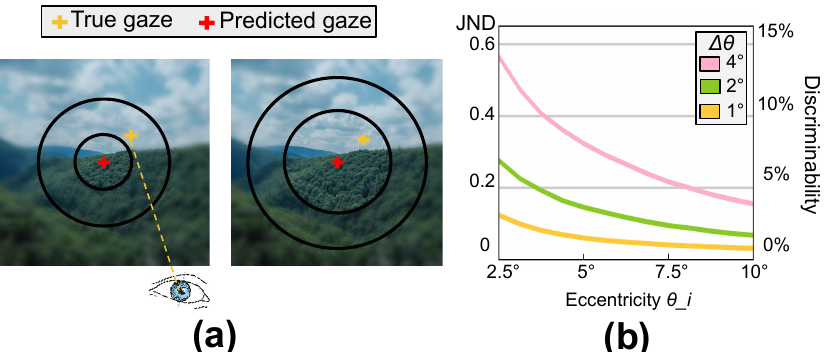}
\caption{(a) Visual quality degradation due to tracking error, and then the foveal region is enlarged for better visual quality. (b) Observer's ability to discriminate foveated image with and without tracking error, measured by JND. $x$-axis indicates the eccentricity angle subtended by the fovea. Left $y$-axis is the JND score, right side of $y$-axis is the discriminability probabilities from ground truth.}
\label{fig:motivation}
\end{figure}

The primary goal of TFR is to reduce computational load while maintaining visual fidelity. To measure the impact on the visual fidelity, traditional visual similarity metrics typically model only image-space context, without considering the observer's visual field \cite{wang2004image,zhang2018perceptual}. 
An emerging body of literature introduces eccentricity angle relative to the fovea into metrics to assess the quality of TFR \cite{chen2024pea,mantiuk2021fovvideovdp,duinkharjav2022image}.
These computational metrics allow for the systematic prediction of foveated image quality, eliminating the need for user studies, which often suffer from limited sample sizes and subjective rating variances.
Following other work in the VR community \cite{huang2023virtual,deng2022fov,bauer2022fovolnet}, we use the FovVideoVDP metric \cite{mantiuk2021fovvideovdp} to assess visual quality in terms of just-noticeable difference (JND)~\cite{cui2022JND}.
Expressing image similarity in terms of JND units offers the advantage of clearly quantifying the probability shift in a population's ability to distinguish between test and reference image pairs, as illustrated in \cref{fig:motivation}(b). 
We observe that, for a fixed eccentricity angle $\theta_{i}$, a larger tracking error $\Delta \theta$ results in a greater increase in discriminability from the ground truth (GT) and a higher JND, indicating poorer foveated image quality.

\subsection{Latency Evaluation of Previous Approaches}
\label{sec:latency_motivation}
While previous work on gaze tracking DNNs has made significant progress in reducing prediction errors, almost all of these approaches focus on minimizing the \textbf{average tracking error} across the training set, resulting in a low average tracking error during execution~\cite{vrpaper_zhu, nvgaze, resnet_inception,mazzeo2021openeds,Model_based_method_1, Model_based_method_2}. 

\begin{figure*}[ht]
\centering
\includegraphics[width=0.91\textwidth]{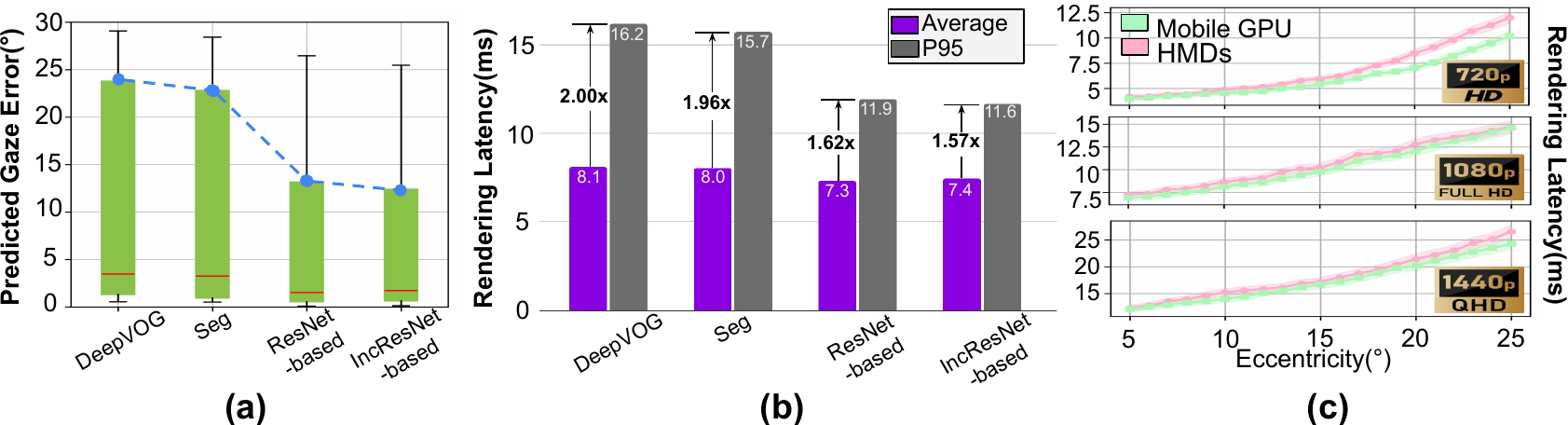}
\caption{(a) Predicted gaze error distributions on the OpenEDS2020 dataset, showing mean, 5th, 95th percentiles, min, and max angular errors. NVgaze results were excluded due to high tracking error and inconsistent performance. (b) Rendering latency for existing methods in both average and max error scenarios with a resolution of $1080\times1920$. (c) Rendering latency increases with eccentricity on HMD and GPU at resolutions $720\times1280$, $1080\times1920$, and $1440\times2560$, with the shaded area showing the $5\%$-$95\%$ confidence interval.}
\label{fig:latency_measurement}
\vspace{-6pt}
\end{figure*}

To study the distribution of the tracking errors $\Delta \theta$ produced by the tracking algorithms, we train and test various gaze tracking DNNs proposed in~\cite{vrpaper_zhu, nvgaze, resnet_inception,mazzeo2021openeds}, on the OpenEDS 2020 dataset~\cite{OpenEDS2020}. \cref{fig:latency_measurement}(a) shows the average tracking errors along with the 95th percentile of the tracking errors on the test dataset of OpenEDS 2020 dataset for five methods: DeepVOG~\cite{deepvog}, Seg~\cite{vrpaper_zhu}, ResNet-based~\cite{mazzeo2021openeds}, IncResNet-based~\cite{resnet_inception} and NVGaze~\cite{nvgaze}. Our observations are as follows: first, all methods, except NVGaze~\cite{nvgaze}, achieve an small average value for $\Delta \theta$, ranging between 2° and 2.5°, with NVGaze showing a higher average tracking error of 9.2°. Notably, the tracking errors exhibit a long-tail distribution, characterized by a high 95th-percentile tracking error, suggesting that these gaze-tracking DNNs may incur large errors for certain eye images.
As a result, the foveated image generated based on the predicted gaze may not align well with what the user is actually looking at, leading to significant degradation in visual quality for the user, as shown in left part of \cref{fig:motivation}(a). As presented in \cref{fig:motivation}(b), a larger error will further lead to a decrease on visual quality.
Unfortunately, most previous work has focused solely on minimizing average gaze tracking performance, and none have optimized gaze tracking DNNs by considering the impact of tracking error distribution in foveated rendering applications.

To maintain visual quality without affecting user experience, the high-resolution foveal and inter-foveal regions must be enlarged to compensate for gaze tracking errors, as shown in right part of \cref{fig:motivation}(a). However, this increases rendering latency, as rendering larger foveal regions at higher resolution raises computational costs.

To investigate how the size of the central foveal region influences the performance of the TFR system, we analyze the changes in rendering latency across different foveal region sizes. Latencies are measured on two types of devices: the Meta Quest Pro~\cite{quest_pro} and the Quadro RTX 3000 Mobile GPU~\cite{mobile_gpu}, which has been utilized by the early work to simulate the behavior of the UMD~\cite{power_quality_tradeoff}. For the Meta Quest Pro, we randomly select 50 frames from the Digital Combat Simulator World (DCS) VR~\cite{DCS} to measure their rendering latency. For mobile GPU, we employ NVIDIA’s Variable Rate Shading (VRS) for rendering and measure the average rendering latency across four scenes, which include a mix of indoor and outdoor environments: Bistro (Outdoor), Sponza (Indoor), Classroom (Indoor), and San-Miguel (Indoor and Outdoor). For both devices, we apply three different rendering resolutions, $720\times 1280$ (720P), $1080\times 1920$ (1080P) and $1440\times 2560$  (1440P). 
 
Following the settings used in previous works~\cite{albert2017latency, nvidia_vrs, foveated_rendering_basic}, we initially set the eccentricity angle \( \theta_{f} \) for the foveal region and inter-foveal regions to 5° and 20°, respectively. Subsequently, we gradually increase \( \theta_{f} \) to 25° and the inter-foveal eccentricity to 40° to account for the gaze tracking errors generated by various gaze tracking DNNs.
The resolution drop of the inter-foveal region and the peripheral regions are set to $4\times$ and $16\times$, respectively. We record the average rendering latencies as the eccentricity angles change.
The results are shown in \cref{fig:latency_measurement}(c). From the data, it is evident that rendering latency increases superlinearly as the eccentricity angle $\theta_{f}$ grows for both the Meta Quest Pro and the Mobile GPU. Notably, the rendering latency nearly doubles when the eccentricity angle reaches 20°, compared to when the eccentricity angle is 5°. Consequently, a large gaze tracking error $\Delta \theta$ will inevitably cause a larger central foveal region to ensure the visual quality, which in turn results in a significant increase in latency overhead. 

To analyze the latency overhead caused by gaze tracking errors for each method, \cref{fig:latency_measurement}(b) shows the rendering latency at a resolution of $1080\times1920$. We set $\Delta \theta$ to represent either the average tracking error or the 95th percentile of the tracking errors for each method, using the test dataset from the OpenEDS 2020 dataset. Our observations reveal that when $\Delta \theta$ is set to the average gaze tracking error, most algorithms achieve very low rendering latency. However, when considering the 95th percentile of the tracking error, all algorithms experience an average of $2.2\times$ increase in tracking error, with some algorithms (such as NVGaze~\cite{nvgaze}, Seg~\cite{vrpaper_zhu}, and ResNet-based~\cite{mazzeo2021openeds}) even exceeding 20 ms, failing to achieve real-time rendering requirement of 60 FPS, according to~\cite{zielinski2015exploring}.

According to \cref{fig:tfr_latency}, the per-frame latency $T_{total}$ for TFR can be expressed as $T_{total} = T_{sensing} + T_{comm} + T_{tracking} + T_{fr}$, where $T_{sensing}, T_{comm}, T_{tracking}, T_{fr}$ represent the latency of camera sensing, MIPI communication, gaze tracking and foveated rendering, respectively. As described in the \cref{sec:latency_breakdown}, $T_{sensing}$ and $T_{comm}$ is small compared with $T_{tracking}$ and $T_{fr}$. We will describe the efficient gaze tracking DNN design to minimize $T_{tracking}$ in \cref{sec:cropping} and \cref{sec:efficient-network}, and discuss a novel training framework of the gaze tracking DNN to minimize $T_{fr}$ in \cref{sec:loss-design}. 

\subsection{Tracking Performance and Latency Tradeoffs in TFR}
In practice, rendering resolution settings for in AR/VR devices can vary significantly during usage, depending on the context and the specific needs of the experience. For example, in highly detailed, visually rich environments like gaming or virtual simulations, users may prefer higher resolutions to capture intricate details and enhance immersion. However, in scenarios requiring rapid interaction or movement, such as fast-paced games or real-time collaboration, users might prioritize performance over maximum resolution to maintain smoothness and responsiveness. 
For popular HMD, such as Apple Vision Pro, supports a render resolution setting as high as $3860\times 3200$~\cite{apple_vision_pro}.
These different resolutions place varying workloads on the host processor, resulting in different rendering times, as presented in \cref{fig:latency_measurement}(c).

A previous study~\cite{power_quality_tradeoff} indicates that the additional implementation overhead associated with executing gaze tracking DNN might surpass the savings achieved through efficient foveated rendering. Specifically, if we denote $T_{full}$ as the execution latency for processing an image frame and rendering it at full resolution, then $T_{tracking}$ could exceed $T_{full} - T_{fr}$. Moreover, since users may dynamically adjust the rendering resolution, $T_{fr}$ can also vary dynamically. Therefore, it is essential to appropriately balance the running costs and tracking accuracy of the gaze tracking DNN to minimize the overall latency.
To ensure that the gaze tracking DNN remains adaptable and performs effectively, we have developed a multi-resolution DNN training approach (\cref{sec:multi-resolution-training}) that simultaneously optimizes numerous sub-networks across a variety of DNN architectures (\cref{fig:multi-resolution-training} (a)). This training method produces a multi-resolution DNN that can produce sub-networks at various resolutions during runtime, enabling selecting the optimal gaze tracking DNN configuration based on the current TFR system conditions.

\section{FovealNet Design}
\label{sec:method}
\begin{figure}[tp]
\centering
\includegraphics[width=0.5\textwidth]{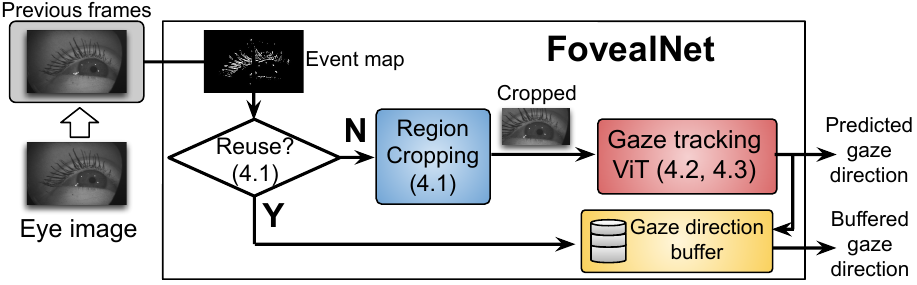}
\caption{Overall architecture of FovealNet.}
\label{fig:overall_foveal}
\end{figure}
The overall structure of FovealNet is illustrated in \cref{fig:overall_foveal}. When an eye image is received, the eye image is cropped using the method outlined in \cref{sec:cropping}. The cropped image is then passed to the gaze tracking ViT, which is explained in \cref{sec:efficient-network} and \cref{sec:loss-design}. Finally, the multi-resolution training approach is described in \cref{sec:multi-resolution-training}.


\SetKwComment{Comment}{/* }{ */}
\SetAlgoLined 

\subsection{Event-based Cropping for Efficient TFR}
\label{sec:cropping}

Eye images captured by near-eye cameras often contain redundant pixels (e.g., background, facial muscles) that are irrelevant for gaze tracking prediction. These pixels can negatively impact the prediction results and increase the computational cost. To tackle this, we propose an event-based analytical approach that efficiently crops the informative regions of the eye. 

\subsubsection{Region Cropping Algorithm}

Given that the pupil is the most relevant area for human gaze, we focus on cropping the informative region of the input frame around the pupil's position. To do this, we employ an efficient analytical approach to precisely detect the pupil location, enabling us to accurately crop a fixed-size region centered on the detected pupil.

\begin{figure}[tp]
\centering
\includegraphics[width = 0.85\linewidth]{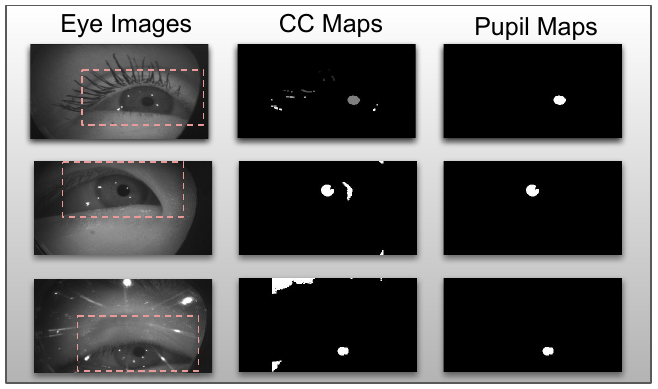}
\caption{
Examples of pupil-centered cropping with intermediate results: connected components (CC) maps and pupil maps. The pink dashed rectangles represent the cropped region.}
\label{figure:event_cropping}
\end{figure}

We begin by applying a masking process to eliminate the background region around the image edges. Then, leveraging the prior knowledge that the pupil is typically darker than the surrounding sclera and iris~\cite{pupil_2013}, we perform inverse binarization to emphasize the darker regions of the image. Next, we apply morphological opening to reduce noise in the image. At this stage, only the pupil and other dark regions like eyelashes remain, as shown in the connected components (CC) maps in \cref{figure:event_cropping}. Given the fact that the pixel density in the pupil region is significantly higher than in other areas. Thus, we can identify the pupil by searching for the largest connected component (LCC) in the image and using the center of this component to represent the pupil's center. Once the LCC is identified in the image, since most pupils are either circular or elliptical~\cite{pupil_2013}, we apply an operation, $is\_pupil$, to determine the shape of the region, by calculating the roundness~\cite{wikipedia_roundness} of this region, if the value indicates an approximate circle or ellipse, we classify it as a pupil.
Once the presence of the pupil is confirmed, a rectangle of predefined size ($450\times 200$) is fitted around the pupil center to crop the informative region. If the rectangle touch the boundary of the image, the rectangle will be translated accordingly (last row of \cref{figure:event_cropping}) to ensure it properly fits within the region.

\subsection{Efficient Gaze Tracking Network}
\label{sec:efficient-network}

The cropped eye images containing informative content are first resized to a smaller square (224×224) and then processed by the gaze tracking DNN to predict gaze direction. FovealNet employs a vision transformer (ViT) architecture~\cite{dosovitskiy2020image} for this task, as it provides superior performance compared to convolutional neural network (CNN)-based architectures. The ViT operates by dividing the input image into patches, which are then tokenized and appended with positional information, the outputs are then passed through the transformer block. The ViT contains 8 transformer block, each block consists of 6 heads with an embedding dimension of 384. We also modify the original ViT by replacing the classifier MLP layers with a sequence of linear layers to output the 2D gaze direction. 

A key advantage of ViT over CNN is its ability to fine-grain prune input tokens, enabling the removal of image tokens with unimportant content, as shown in \cref{fig:token-pruning}. In the self-attention mechanism, tokens are linearly transformed into Query, Key, and Value matrices. The attention score is then computed by performing a dot product between the Query and Key matrices, followed by scaling and Softmax operation. The attention score reflects the importance of each token in relation to the gaze prediction result. 

\subsection{Performance-aware Training Strategy}
\label{sec:loss-design}
As discussed in Section~\ref{sec:latency_motivation}, most existing gaze-tracking DNNs focus on minimizing the average tracking error. However, this often leads to a higher 95th percentile error in gaze tracking $\Delta \theta$, which increases rendering times in the TFR system. We propose a training strategy to  addresses this issue by minimizing the maximum tracking error during training, which can be formulated as:
\begin{equation}
\label{eqn:minmax-original}
\min \max_{d \in D_{train}}(||\theta_{d} - \theta^{g}_{d}||^{2})
\end{equation}
where $\theta_{d}$ and $\theta^{g}_{d}$ denote the predicted gaze direction and the ground-truth gaze direction (in radians) for the input sample $d$ in the training dataset $D_{train}$, respectively. To enhance training stability, the DNN is trained using multiple batches of training samples, resulting in \cref{eqn:minmax} being:
\begin{equation}
\label{eqn:minmax}
\min \sum_{b\in B}\max_{d \in D^{b}_{train}}(||\theta_{d} - \theta^{g}_{d}||^{2})
\end{equation}
where $B$ denotes the set of training dataset batches, and $D^{b}_{train}$ represents the set of training data in batch b. However, using this formula directly as the loss function can result in underutilization of the training dataset, as it tends to focus on only optimizing the sample with the highest tracking error. Empirically, we find it more effective to optimize an approximate version of \cref{eqn:minmax} by replacing the max operation with an alternative approach, using the approximation $max(x_1, x_2) \approx \frac{1}{N}\ln(e^{Nx_{1}} + e^{Nx_{2}})$, namely:
\begin{equation}
  \label{eqn:softmax}
  \sum_{b\in B} \frac{1}{N} \ln\left(\sum_{d \in D^{b}_{train}} e^{N||\theta_{d} - \theta^{g}_{d}||^{2}} \right)  
\end{equation}
where $N$ is the scaling factor that controls the temperature of the approximation. During the training process, the value of $N$ is tuned carefully to adapt to the value distribution of the input training data to ensure the better convergence of the training process.

Finally, we can directly relate the gaze error from \cref{eqn:softmax} to the TFR latency, enabling us to optimize $T_{tf}$. To achieve this, we profile rendering latencies across different VR devices and develop a piecewise linear function $U(.)$ that links the gaze tracking error to the corresponding rendering latency, as shown in \cref{fig:tfr_latency}(a), facilitating the minimization of rendering overhead while preserving visual quality. Thus, the training objective becomes:
\begin{equation}
    \label{eqn:final-loss}
    \sum_{b\in B} U\left(\frac{1}{N} \ln\left( \sum_{d \in D^{b}_{train}} e^{N||\theta_{d} - \theta^{g}_{d}||^{2}} \right) \right)
\end{equation}

\subsection{Multi-resolution Training Mechanism}
\label{sec:multi-resolution-training}
\begin{figure}[t]
\centering
\includegraphics[width=0.95\linewidth]{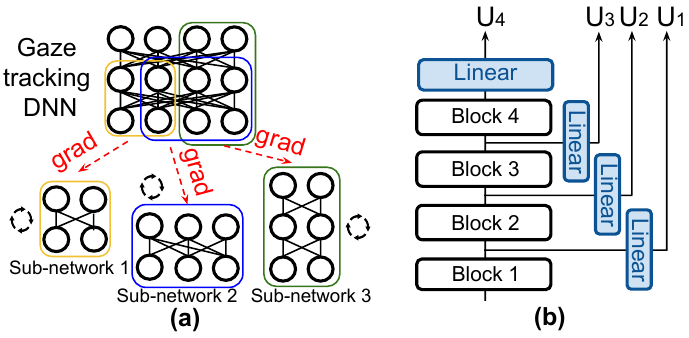}
\caption{(a) Multi-resolution training framework. (b) An example of four layer ViT with early-exit mechanism, the output from each early-exit is sent to a linear layer to produce the gaze prediction.}
\label{fig:multi-resolution-training}
\vspace{-10pt}
\end{figure}
In \cref{sec:loss-design}, we depict the design of the loss function (\cref{eqn:final-loss}) by directly optimizing the rendering latency. 
In practice, the processing latency $T_{tracking}$ of gaze tracking DNN will also contribute to the total latency $T_{total}$, and the hardware processing latencies for rendering and tracking can vary due to user settings and resource sharing with other applications.

To maintain adaptability and performance, we develop a multi-resolution DNN training approach that optimizes multiple subnetworks across various DNN architectures simultaneously (\cref{fig:multi-resolution-training} (a)). This joint-optimization training framework will produce a multi-resolution model that can execute at varying depths, allowing for the selection of the optimal gaze tracking DNN based on the current system conditions.

To achieve this, we attach a linear layer at the end of each encoder block within the gaze tracking ViT, which will produce a prediction on gaze direction based on the intermediate results from the early-exit points, as shown in \cref{fig:multi-resolution-training} (b).
Specifically, let $L$ denote the total number of layer blocks within the ViT, and $U_{l}$ denote the loss generated from the output of layer $l\in L$. The loss function $U_{multi}$ for the multiresolution training can be formalized as:
\begin{equation}
    \label{eqn:multiresolution-loss}
    U_{multi} = \sum_{l} U_{l}
\end{equation}

Using early-exit mechanisms, the resulting gaze tracking DNN can operate at different depths to balance gaze tracking latency and accuracy. During execution, the depth can be adjusted adaptively based on the TFR system’s condition to optimize the overall per-frame latency $T_{total}$.

\begin{table}[tp]
\centering
\caption{Accuracy performance and computational cost of different approaches by minimizing average tracking error. P0 and P95 represent $90\%$ and $95\%$ percentiles, respectively.}
\resizebox{\linewidth}{!}{%
\begin{tabular}{c|ccccc|cc}
\toprule
Network & Mean & P90 & P95 & Min & Max & FLOPS (billions)\\
\midrule
NVGaze~\cite{nvgaze} & 6.81 & 13.07 & 18.62 & 0.94 & 42.30 & 0.021\\
DeepVoG~\cite{deepvog} & 3.47 & 17.76 & 23.77 & 0.55 & 29.06 & 36.5\\
Seg~\cite{vrpaper_zhu} & 3.25 & 18.29 & 22.80 & 0.52 & 28.42 & 2.6 \\
ResNet-based~\cite{mazzeo2021openeds} & 1.52 & 5.96 & 13.15 & 0.07 & 26.46 & 3.6 \\
IncResNet-based~\cite{resnet_inception} & 1.72 & 6.23 & 12.4 & 0.12 & 25.47 & 13.12\\
FovealNet (0.2) & 1.27 & 4.92 & \textbf{8.09} & \textbf{0} & 24.92 & 2.08\\
FovealNet (0.1) & 1.05 & 5.75 & 9.63 & \textbf{0} & 25.54 & 2.42\\
FovealNet (0.0) & \textbf{0.93} & \textbf{4.71} & 8.21 & \textbf{0} & \textbf{24.2} & 2.80 \\
\bottomrule
\end{tabular}%
}
\label{table:results_minavg}
\end{table}
\section{Tracking Performance Evaluation}
\label{sec:accuracy-evaluation}
\subsection{Settings}
\label{sec:evaluation-setting}
We evaluate the tracking performance of FovealNet using the OpenEDS2020 dataset~\cite{OpenEDS2020}, which consists of 128,000 images from 32 participants in the training set and 70,400 images from 8 participants in the validation set. All participants wore a VR-HMD, and the images were captured with a near-eye camera operating at 100Hz and a resolution of $640\times400$ pixels. The dataset includes ground truth 3D gaze vectors, which we converted into 2D gaze vectors (horizontal and vertical components)~\cite{appearance_zhang_1} to enable a more effective evaluation.

To evaluate FovealNet, we select two model-based methods and three appearance-based methods. The model-based approaches include Seg~\cite{vrpaper_zhu}, an efficiency-focused segmentation network, and DeepVoG~\cite{deepvog}, a popular encoder-decoder network for gaze tracking. The appearance-based approaches are NVGaze~\cite{nvgaze}, a one-shot CNN-based network, and ResNet-based~\cite{mazzeo2021openeds} and Inception-ResNet-based models~\cite{resnet_inception}. For the model-based DNNs, we download the pre-trained models from the code repository~\cite{code_github} and run the code using the settings reported in their respective papers.
However, since no pre-trained weights are available for the appearance-based methods, we train these methods using the same procedure reported in their work, and report the corresponding results.

In the data pre-processing stage, we first horizontally flip the right eye images to align them with the left eye images, a common practice that allows the application of a single mapping for both eyes~\cite{appearance_zhang_1}. We apply $\beta_1 = 0.2$ and $\beta_2 = 500$ for cropping the input image in \cref{eqn:cropping}. To train the gaze tracking ViT, we use a series of data augmentations techniques to enhance the training convergence speed. These techniques include random cropping, where the image is cropped to a random size between $80\%$ to $100\%$ of its original size, followed by a random shift in position by up to $10\%$ of the image’s width and height. Finally, we normalize the images to standardize the input data for the model. 

To train the FovealNet, we utilize the Adam optimizer with a learning rate of 5e-4 and a momentum of 0.9, a step learning scheduler that reduces the learning rate by a factor of 0.2 every 10 epochs and a batch size of 512 for 100 epochs. We also implement an early stopping mechanism that halts training if no improvement is observed on the validation set after 10 consecutive epochs.  Specifically, we utilize the Adam optimizer with a learning rate of 5e-5 and a momentum of 0.9 and a batch size of 512 for 50 epochs to finetune the token pruned FovealNet. All experiments are conducted on a single RTX 4090D GPU.

We evaluate the performance of FovealNet by comparing it with other baseline approaches under two training objectives. First, we train FovealNet using an objective function that aims to minimize the average gaze tracking error, which is the same training objective used by all other baselines. We show that even with this baseline objective, FovealNet still outperforms all other approaches (\cref{sec:min-avg-performance}). Next, we train FovealNet using the objective function described in \cref{eqn:final-loss} and compare its performance against other methods in \cref{sec:min-max-performance}. Finally, we present the performance of FovealNet at different resolutions in \cref{sec:multi-resolution-accuracy}.

\subsection{Accuracy Evaluation by Minimizing Average Gaze Error}
\label{sec:min-avg-performance}
We train all the models using an objective function designed to minimize the average gaze tracking error, defined as: $L_{mse} = \sum_{b\in B}\sum_{d\in D_{train}^{b}} (y_d - \hat{y}_d)^2$. 
B denotes the number of batches within the training data, $y_{d}$ and $\hat{y}_d$ represent the predicted and ground truth gaze direction for training data $d$. This is the same training objective used by all other previous works on gaze tracking.
For DeepVoG~\cite{deepvog} and Seg~\cite{vrpaper_zhu}, we use their pretrained model weights and deploy them directly for evaluation. For FovealNet, we evaluate its performance under three settings with tokenwise pruning ratios of 0.2, 0.1, and 0.0 (no pruning), labeled as FovealNet (0.2), FovealNet (0.1), and FovealNet (0.0), respectively.

\cref{table:results_minavg} compares the predicted gaze error of our algorithms against various baselines. To represent the predicted gaze error, we use the eccentricity from true gaze direction, a method that widely utilized in~\cite{mazzeo2021openeds, nvgaze}. FovealNet (0.0), without the token pruning mechanism, maintains a minimal mean gaze error of \(0.93^\circ\) across all evaluated models. When different token pruning ratios are applied, the models exhibit gaze errors of \(1.05^\circ\) and \(1.27^\circ\), respectively, both of which are lower than those of most baseline methods. Additionally, our models show the lowest values for both the 95th-percentile and maximum errors. 
Specifically, the FovealNet (0.2) achieves the smallest 95th-percentile error of $8.09^\circ$.

We also compare the computational complexity in terms of Floating Point Operations (FLOPs), as shown in \cref{table:results_minavg}. FovealNet (0.2) and FovealNet (0.1) achieve lower computational costs compared to most methods, except for NVGaze, which has significantly worse tracking performance. Specifically, the computational cost of FovealNet (0.2) is about $15\%$ of that of the IncResNet-based model~\cite{resnet_inception}, $70\%$ lower than the ResNet-based model~\cite{mazzeo2021openeds}, and $27\%$ lower comparable to Seg~\cite{vrpaper_zhu}.



\subsection{Accuracy Evaluation by Performance-aware Loss}
\label{sec:min-max-performance}
\begin{figure}[tp]
\centering
\includegraphics[width=1\linewidth]{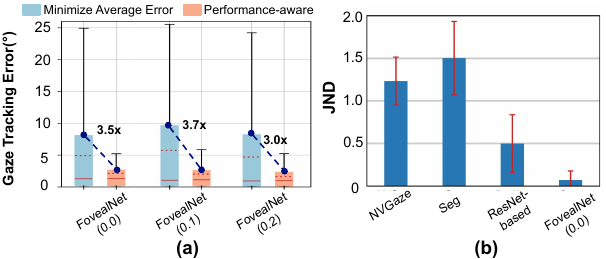}
\caption{(a) Distribution of gaze tracking error of FovealNet trained with performance-aware loss. The box plot covers 5th to 95th percentile of gaze tracking error, red line denotes average error, red dashed line denotes 90th percentile error. (b) Perceptual Quality Measurement in JND.}
\label{fig:gaze_minmax}
\end{figure}

In this section, we train FovealNet using the performance-aware loss function specified in \cref{eqn:final-loss} with a scaling factor $N$ of 100. Specifically, we use the rendering latency measurements of the Meta Quest Pro on input images with a size of $1080 \times 1920$, as shown in \cref{fig:latency_measurement}(c), to generate the piecewise linear function $U(.)$ and evaluate its impact on the gaze tracking error for FovealNet.

\cref{fig:gaze_minmax} shows the effectiveness of the performance-aware training strategy from \cref{sec:loss-design}. The blue bars indicate the error distribution using the objective function that minimizes average gaze error, as in \cref{table:results_minavg}, while the orange bars represent the gaze error distribution using the performance-aware loss function.

For FovealNet (0.0), it exhibits a notable reduction in the 95th-percentile tracking error, decreasing from 8.21° to 2.31°, with the 90th-percentile error reducing from 4.71° to 1.62° and maximum error diminishing significantly from 24.2° to 5.22°. For FovealNet (0.1) and FovealNet (0.2), the pruned models show only a minor accuracy degradation, with a 90th-percentile error of 2.02° and 2.13° and 95th-percentile error of 2.6° and 2.72°. Overall, by using the performance-aware training strategy, our model achieves an average reduction of over $65\%$ in 95th-percentile and over $70\%$ in maximum errors compared to minimizing average error strategy. Finally, compared to other baselines that aim to minimize the average gaze tracking error shown in \cref{table:results_minavg}, the performance-aware training strategies significantly improve the worst-case gaze error distribution, leading to notable system performance enhancements, as detailed in \cref{sec:system-evaluation}.

\begin{table}[tp]
\centering
\caption{Evaluation of Multi-resolution FovealNet.}
\resizebox{0.85\linewidth}{!}{%
\begin{tabular}{c|cccc|c}
\toprule
\multirow{2}{*}{Model Depth} & \multicolumn{4}{c|}{{Gaze Error (°)}} &  \multirow{2}{*}{FLOPs (Billion) }\\
\cmidrule(lr){2-5}
 & Mean & P90 & P95 & Max &  \\
\midrule
3 & 2.93 &5.23 & 7.35 & 15.70 & 1.06 \\
4 & 1.90 &3.87 & 5.23 & 10.36 & 1.41 \\
5 & 1.68 &3.43 & 3.98 & 8.28 & 1.76 \\
6 & 1.30 &2.78 & 3.38 & 7.78 & 2.10 \\
7 & 1.15 &2.61 & 3.05 & 6.98 & 2.45 \\
8 & 1.08 &1.95 & 2.54 & 5.94 & 2.80 \\
\bottomrule
\end{tabular}%
}
\label{table:multiresolution}
\vspace{-8pt}
\end{table}
\subsection{Perceptual Quality Measurement}
\label{sec:vdp}
To evaluate the impact of gaze-tracking errors on the perceptual quality of the foveated output, we use the FovVideoVDP metric in \cref{sec:fvvdp}.
Specifically, we sampled $400$ random images from the MS COCO test dataset~\cite{lin2014microsoft}, and applied the foveation algorithm~\cite{perry2002gaze,jiang2015salicon}, configured with a  $\theta_{i} = 5^\circ$ of the eccentricity angle subtended by fovea, when displayed on an HTC Vive Pro HMD (i.e., $13.2^\circ$ pixels per degree~\cite{mantiuk2021fovvideovdp}).

For each image, $1080$ gaze snapshots were used to simulate foveated rendering corresponding to each snapshot.
We measure the similarity between images generated with the predicted gaze direction and those generated with the ground truth gaze direction for each gaze snapshot. We adopt the FovealNet(0.0) discussed in~\cref{sec:min-max-performance} for evaluation.
\cref{fig:gaze_minmax}(b) shows the experiment results. Our FovealNet(0.0) achieves a minimal JND of 0.07, meaning users were unable to perceive any noticeable difference from the ground truth. In comparison, the ResNet-based model produces a JND of 0.5, corresponding to a 13\% increase in discriminability, while the Seg model results in a JND of 1.5, indicating a 34\% likelihood that the rendered image is significantly distinguishable from the ground truth. Statistical analysis reveals a highly significant difference in visual quality between our method and the others, with a p-value of less than $10^{-6}$, suggesting the observed differences are extremely unlikely to be due to chance.
\begin{figure*}[tp]
\centering
\includegraphics[width=1\linewidth]{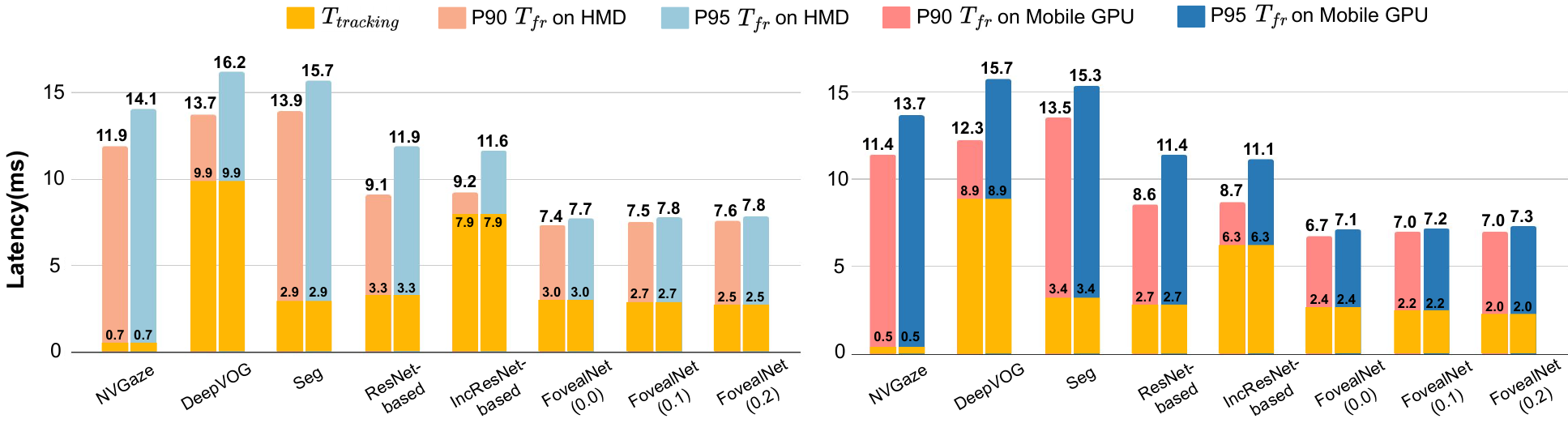}
\caption{Evaluation on overall processing latency. (a) Measurement on Quest Pro at a resolution of 1080P (1080$\times$1920), including 90th-percentile rendering latency, 95th-percentile rendering latency and tracking latency. (b) Measurement on mobile GPU at a resolution of 1080P (1080$\times$1920), including 90th-percentile rendering latency, 95th-percentile rendering latency and tracking latency.}
\label{fig:minmax_tfr}
\vspace{-5pt}
\end{figure*}


\subsection{Multi-Resolution Accuracy}
\label{sec:multi-resolution-accuracy}
While \cref{sec:min-avg-performance} and \cref{sec:min-max-performance} focus on single-resolution training with the performance-aware training strategy, this section evaluates the performance of FovealNet using the multi-resolution training method from \cref{sec:multi-resolution-training}. 
Specifically, six early exits branches are introduced at the end of each transformer block of gaze tracking ViT from block 3 to 8, with a small linear layer for gaze prediction at each exit, resulting in six subnetworks of varying depths. During training, we compute the training losses from the six loss functions and sum them to generate the final multi-resolution loss, as shown in \cref{eqn:multiresolution-loss}. The rest of the training settings follows \cref{sec:min-max-performance}.

\cref{table:multiresolution} shows the gaze tracking error and computational cost for each sub-network. FovealNet with a depth of 3 reduces computational complexity by $62.1\%$ compared to the 8-layer subnetwork, while achieving a 95th-percentile error of 7.35° and a maximum error of 15.70°. Notably, the performance 8-layer subnetwork shows a 0.1° increase in mean error, a 0.33° increase in 90th-percentile error and a 0.23° increase in 95th-percentile error compared to the 8-layer FovealNet trained solely with the loss at layer 8. This increase is because of the changes in the training loss function from a single loss (\cref{eqn:final-loss}) to multi-resolution loss (\cref{eqn:multiresolution-loss}).


\subsection{Ablation Study}
\label{sec:ablation}
In this section, we first examine the effect of the input cropping method, introduced in \cref{sec:cropping}, on the gaze tracking performance and computational cost of FovealNet. \cref{table:ablation} presents the tracking errors and computational cost of FovealNet when using cropped versus uncropped input. For the gaze tracking DNN to process the uncropped input, it is also resized to a square shape of 224×224 pixels.
We observe that the involvement of the cropping method results in a mean error decrease from 1.1° to 0.98°, and 90th-percentile and 95th-percentile error drops 0.2° and 0.34° respectively. Meanwhile, the introduction of cropping only increase 11M FLOPs, results in a $<0.5\%$ FLOPs increase.

\begin{table}[tp]
\centering
\caption{Performance with and without cropping method (in degrees).}
\resizebox{\linewidth}{!}{%
\begin{tabular}{c|ccccc}
\toprule
Method & Mean error & P90 error & P95 error & Max error & FLOPs (Billions) \\
\midrule
with cropping & 0.98 & 1.62 & 2.31 & 5.23 &2.810\\
with/o cropping & 1.10 & 1.82 & 2.65 &5.98	&2.799\\
\bottomrule
\end{tabular}%
}
\label{table:ablation}
\vspace{-5pt}
\end{table}

In \cref{sec:min-max-performance}, when training with the performance-aware loss, we use a scaling factor of $N = 100$. Here, we explore the impact of different choices of the scaling factor $N$. As shown in \cref{table:ablation-N}, for a smaller value of $N = 10$, \cref{eqn:softmax} is unable to effectively minimize the maximum error, leading to a large 95th-percentile error of 6.92°. On the other hand, excessively large values of $N$, such as 150 or 200, may cause overflow during training, resulting in training failure. Thus, selecting an appropriate $N$ is crucial for ensuring the effectiveness of \cref{eqn:final-loss}.

\begin{table}[tp]
\centering
\caption{Performance with different scaling factor $N$ (in degrees).}
\resizebox{0.7\linewidth}{!}{%
\begin{tabular}{c|cccc}
\toprule
$N$ & Mean & P90 error & P95 error & Max error\\
\midrule
10 & 0.97 & 4.57 & 6.92 & 18.98\\
50 & 1.02 & 1.68 & 2.42 & 5.37 \\
100 & 0.98 & 1.62 & 2.31 & 5.23\\
150 & Inf & Inf & Inf &Inf \\
200 & Inf & Inf & Inf &Inf \\
\bottomrule
\end{tabular}%
}
\label{table:ablation-N}
\vspace{-10pt}
\end{table}


\section{TFR System Performance Evaluation}
\label{sec:system-evaluation}
In this section, we evaluate the system performance by measuring processing latency for various methods. 
First, in \cref{sec:system-performance-aware}, we evaluate the system performance of the FovealNet trained with the performance-aware loss defined in ~\cref{eqn:final-loss}. We then demonstrate the system performance of FovealNet under different rendering system conditions by switching the rendering resolution in \cref{sec:multi-resolution-performance}.

\subsection{Evaluation with Performance-aware Training Loss}
\label{sec:system-performance-aware}
In this section, we compare the system performance across different approaches with the performance-aware training of FovealNet, considering both $T_{tracking}$ and various $T_{fr}$ values. For $T_{fr}$, we evaluate the foveated rendering configurations that account for the 90th-percentile and 95th-percentile gaze tracking errors to ensure greater versatility. The $T_{fr}$ values are derived by determining latency based on the latency analysis outlined in \cref{sec:latency_motivation} based on gaze tracking error. We compare the latency performance of various approaches on both HMD (Meta Quest Pro) and a mobile GPU (Quadro RTX 3000). For FovealNet, we train it using the performance-aware loss function described in \cref{eqn:final-loss}, where $U(.)$ represents the processing latency under different eccentricity angle $\theta_{f}$ on either the HMD or mobile GPU. To compute the $\theta_{f}$, we set the eccentricity angle $\theta_{i}$ subtended by the fovea to 5°, and $\Delta \theta$ is set to P95 or P90 of the gaze error distribution on OpenEDS 2020 for different approaches. We adopt the single resolution FovealNet that contains 8 ViT blocks under different pruning ratios.

We profile the processing latency $T_{tracking}$ of FovealNet on both the Meta Quest Pro and a Quadro RTX 3000 Mobile GPU, as discussed in \cref{sec:motivation}. Since we do not have access to run ViT on the Meta Quest Pro directly, we use GPGPU-sim~\cite{gpusim} to simulate the performance of the Adreno 650, which is integrated into the Qualcomm Snapdragon XR2+ and deployed in the Meta Quest Pro. The GPGPU-sim is configured according to the specifications of the Adreno 650~\cite{adreno}.

As shown in \cref{fig:minmax_tfr}, for HMD, FovealNet (0.0) achieves the lowest $T_{fr}$ values of 7.4ms and 7.7ms when setting $\Delta \theta$ to P95 or P90 of the gaze error distribution, representing a reduction of at least 1.7ms and 3.9ms compared to previous methods. Since \( T_{sensing} \) and \( T_{comm} \) are relatively small and consistent across different methods, as shown in \cref{fig:gaze_tracking}, the per frame latency \( T_{total} \) will be mainly determined by \( T_{tracking} + T_{fr} \).
 Our FovealNet (0.0) achieves a $T_{tracking} + T_{fr}$ values of 10.4ms and 10.7ms for P90 and P95, respectively. For the FovealNet with tokenwise pruning, the slight growth in gaze prediction error results in only a minimal increase in $T_{fr}$, staying below 0.3ms when compared to the FovealNet (0.0). Notably, FovealNet (0.2) achieves the lowest $T_{tracking} + T_{fr}$, with 10.2ms in P90 scenario and 10.4ms in P95 scenario. Similarly in the evaluation results on mobile GPU, FovealNet (0.0) achieves the lowest $T_{fr}$ across different scenarios. This demonstrates the effectiveness of our performance-aware training strategy.



\subsection{Latency Evaluation under Varying Conditions}
\label{sec:multi-resolution-performance}
\begin{figure}[h]
\centering
\includegraphics[width=1\linewidth]{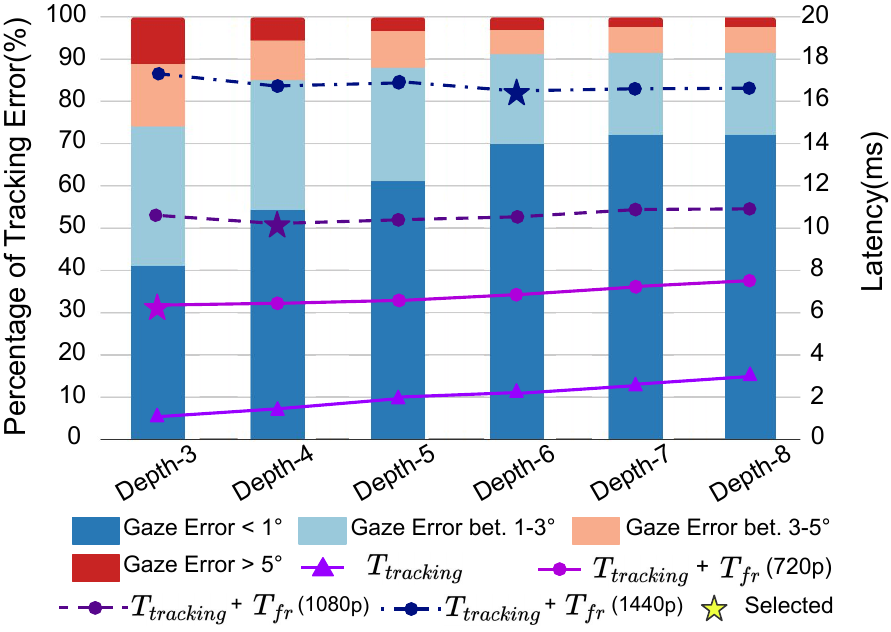}
\caption{The stacked bars show the distribution of the gaze errors for each subnetwork of FovealNet. The lines show the $T_{tracking}$ under different depth and $T_{tracking} + T_{fr}$ under different rendering resolution. The star marks the optimal selection of the subnetwork that achieves the lowest $T_{tracking} + T_{fr}$ at each rendering resolution.}
\label{fig:multiresolution_measurment}
\vspace{-10pt}
\end{figure}

In this section, we evaluate the performance of the multi-resolution FovealNet by using the same model described in \cref{sec:multi-resolution-accuracy} and simulating the corresponding execution latencies on the Meta Quest Pro, following the methods outlined in \cref{sec:system-performance-aware}. We simulate variations in system configuration by adjusting the rendering resolution to 720P ($720\times1080$), 1080P ($1080\times1920$), and 1440P ($1440\times2560$). For each rendering resolution, we assess the $T_{tracking}$ and $T_{fr}$ of the FovealNet at different depths, as produced by the multi-resolution training mechanism detailed in \cref{sec:multi-resolution-training}, and search for the optimal FovealNet configurations to minimize the per-frame latency $T_{total}$. We assume that the TFR system is capable of adapting its optimal rendering settings by adjusting the eccentricity angles \( \theta_{f} \). 

The results are shown in \cref{fig:multiresolution_measurment}. 
The distribution of gaze error and $T_{tracking}$ across the sub-networks remains consistent, with the lowest $T_{tracking}$ at 1.14 ms and the highest at 3 ms. However, in low-resolution 720P scenarios, increasing the depth of the sub-networks offers minimal reduction in $T_{fr}$ while introducing additional $T_{tracking}$. With a sub-network depth of 3, we achieve the optimal configuration, resulting in a $T_{tracking} + T_{fr}$ of 6.19 ms. In contrast, for high-resolution 1440P settings, deepening the sub-networks significantly reduces $T_{fr}$. The optimal configuration is found at a depth of 6, yielding the lowest latency of 16.4 ms.

\section{Conclusion}
\label{sec:conclusion}
In this work, we introduce FovealNet, an AI-based gaze tracking solution designed to enhance the performance of TFR systems. FovealNet can be directly optimized using a loss function that incorporates system performance metrics, resulting in superior outcomes compared to baseline algorithms. To further reduce the implementation cost of the gaze tracking algorithm, FovealNet utilizes an event-based cropping technique that discards irrelevant pixels from the input image. Moreover, it features an efficient token-pruning strategy that dynamically eliminates tokens during processing without sacrificing tracking accuracy.

\newpage
\bibliographystyle{unsrt}
\bibliography{reference} %

\begin{thebibliography}{10}

\bibitem{patney2016perceptually}
Anjul Patney, Joohwan Kim, Marco Salvi, Anton Kaplanyan, Chris Wyman, Nir Benty, Aaron Lefohn, and David Luebke.
\newblock Perceptually-based foveated virtual reality.
\newblock In {\em ACM SIGGRAPH 2016 Emerging Technologies}, SIGGRAPH '16, New York, NY, USA, 2016. Association for Computing Machinery.

\bibitem{albert2017latency}
Rachel Albert, Anjul Patney, David Luebke, and Joohwan Kim.
\newblock Latency requirements for foveated rendering in virtual reality.
\newblock {\em ACM Transactions on Applied Perception (TAP)}, 14(4):1--13, 2017.

\bibitem{franke2021time}
Linus Franke, Laura Fink, Jana Martschinke, Kai Selgrad, and Marc Stamminger.
\newblock Time-warped foveated rendering for virtual reality headsets.
\newblock In {\em Computer Graphics Forum}, volume~40, pages 110--123. Wiley Online Library, 2021.

\bibitem{kaplanyan2019deepfovea}
Anton~S Kaplanyan, Anton Sochenov, Thomas Leimk{\"u}hler, Mikhail Okunev, Todd Goodall, and Gizem Rufo.
\newblock Deepfovea: Neural reconstruction for foveated rendering and video compression using learned statistics of natural videos.
\newblock {\em ACM Transactions on Graphics (TOG)}, 38(6):1--13, 2019.

\bibitem{illahi2020interplay}
Gazi~Karam Illahi, Matti Siekkinen, Teemu K{\"a}m{\"a}r{\"a}inen, and Antti Yl{\"a}-J{\"a}{\"a}ski.
\newblock On the interplay of foveated rendering and video encoding.
\newblock In {\em Proceedings of the 26th ACM Symposium on Virtual Reality Software and Technology}, pages 1--3, 2020.

\bibitem{illahi2021foveated}
Gazi~Karam Illahi, Matti Siekkinen, Teemu K{\"a}m{\"a}r{\"a}inen, and Antti Yl{\"a}-J{\"a}{\"a}ski.
\newblock Foveated streaming of real-time graphics.
\newblock In {\em Proceedings of the 12th ACM Multimedia Systems Conference}, pages 214--226, 2021.

\bibitem{hegazy2019content}
Mohamed Hegazy, Khaled Diab, Mehdi Saeedi, Boris Ivanovic, Ihab Amer, Yang Liu, Gabor Sines, and Mohamed Hefeeda.
\newblock Content-aware video encoding for cloud gaming.
\newblock In {\em Proceedings of the 10th ACM multimedia systems conference}, pages 60--73, 2019.

\bibitem{illahi2020cloud}
Gazi~Karam Illahi, Thomas~Van Gemert, Matti Siekkinen, Enrico Masala, Antti Oulasvirta, and Antti Yl{\"a}-J{\"a}{\"a}ski.
\newblock Cloud gaming with foveated video encoding.
\newblock {\em ACM Transactions on Multimedia Computing, Communications, and Applications (TOMM)}, 16:1--24, 2020.

\bibitem{zou2021enhancing}
Wenjie Zou, Shixuan Feng, Xionghui Mao, Fuzheng Yang, and Zhibin Ma.
\newblock Enhancing quality of experience for cloud virtual reality gaming: An object-aware video encoding.
\newblock In {\em 2021 IEEE International Conference on Multimedia \& Expo Workshops (ICMEW)}, pages 1--6. IEEE, 2021.

\bibitem{deepvog}
Yuk-Hoi Yiu, Moustafa Aboulatta, Theresa Raiser, Leoni Ophey, Virginia~L. Flanagin, Peter {zu Eulenburg}, and Seyed-Ahmad Ahmadi.
\newblock Deepvog: Open-source pupil segmentation and gaze estimation in neuroscience using deep learning.
\newblock {\em Journal of Neuroscience Methods}, 324:108307, 2019.

\bibitem{vrpaper_zhu}
Yu~Feng, Nathan Goulding-Hotta, Asif Khan, Hans Reyserhove, and Yuhao Zhu.
\newblock Real-time gaze tracking with event-driven eye segmentation.
\newblock In {\em 2022 IEEE Conference on Virtual Reality and 3D User Interfaces (VR)}, pages 399--408, 2022.

\bibitem{etracker}
Bin Li, Hong Fu, Desheng Wen, and WaiLun LO.
\newblock Etracker: A mobile gaze-tracking system with near-eye display based on a combined gaze-tracking algorithm.
\newblock {\em Sensors}, 18(5), 2018.

\bibitem{angelo2021vrpaper}
Anastasios~N. Angelopoulos, Julien~N.P. Martel, Amit~P. Kohli, Jörg Conradt, and Gordon Wetzstein.
\newblock Event-based near-eye gaze tracking beyond 10,000 hz.
\newblock {\em IEEE Transactions on Visualization and Computer Graphics}, 27(5):2577--2586, 2021.

\bibitem{Kothari2020EllSegAE}
Rakshit Kothari, Aayush~Kumar Chaudhary, Reynold~J. Bailey, Jeff~B. Pelz, and Gabriel~J. Diaz.
\newblock Ellseg: An ellipse segmentation framework for robust gaze tracking.
\newblock {\em IEEE Transactions on Visualization and Computer Graphics}, 27:2757--2767, 2020.

\bibitem{Overview_tracking}
Dan~Witzner Hansen and Qiang Ji.
\newblock In the eye of the beholder: A survey of models for eyes and gaze.
\newblock {\em IEEE Transactions on Pattern Analysis and Machine Intelligence}, 32(3):478--500, 2010.

\bibitem{Over_tracking_1}
Xucong Zhang, Yusuke Sugano, and Andreas Bulling.
\newblock Evaluation of appearance-based methods and implications for gaze-based applications.
\newblock In {\em Proceedings of the 2019 CHI Conference on Human Factors in Computing Systems}. ACM, May 2019.

\bibitem{Model_based_method}
Lech \'Swirski and Neil~A. Dodgson.
\newblock A fully-automatic, temporal approach to single camera, glint-free 3d eye model fitting [abstract].
\newblock In {\em Proceedings of ECEM 2013}, August 2013.

\bibitem{Model_based_method_1}
Kang Wang and Qiang Ji.
\newblock Real time eye gaze tracking with 3d deformable eye-face model.
\newblock In {\em 2017 IEEE International Conference on Computer Vision (ICCV)}, pages 1003--1011, 2017.

\bibitem{Model_based_method_2}
Erroll Wood, Tadas Baltru{\v{s}}aitis, Louis-Philippe Morency, Peter Robinson, and Andreas Bulling.
\newblock A 3d morphable eye region model for gaze estimation.
\newblock In Bastian Leibe, Jiri Matas, Nicu Sebe, and Max Welling, editors, {\em Computer Vision -- ECCV 2016}, pages 297--313, Cham, 2016. Springer International Publishing.

\bibitem{lu2022model}
Conny Lu, Praneeth Chakravarthula, Kaihao Liu, Xixiang Liu, Siyuan Li, and Henry Fuchs.
\newblock Neural 3d gaze: 3d pupil localization and gaze tracking based on anatomical eye model and neural refraction correction.
\newblock In {\em 2022 IEEE International Symposium on Mixed and Augmented Reality (ISMAR)}, pages 375--383, 2022.

\bibitem{zhang2024swifteye}
Tongyu Zhang, Yiran Shen, Guangrong Zhao, Lin Wang, Xiaoming Chen, Lu~Bai, and Yuanfeng Zhou.
\newblock Swift-eye: Towards anti-blink pupil tracking for precise and robust high-frequency near-eye movement analysis with event cameras.
\newblock {\em IEEE Transactions on Visualization and Computer Graphics}, 30(5):2077--2086, 2024.

\bibitem{RITnet}
Aayush~K. Chaudhary, Rakshit Kothari, Manoj Acharya, Shusil Dangi, Nitinraj Nair, Reynold Bailey, Christopher Kanan, Gabriel Diaz, and Jeff~B. Pelz.
\newblock Ritnet: Real-time semantic segmentation of the eye for gaze tracking.
\newblock In {\em 2019 IEEE/CVF International Conference on Computer Vision Workshop (ICCVW)}. IEEE, 2019.

\bibitem{appearance_zhang_1}
Xucong Zhang, Yusuke Sugano, Mario Fritz, and Andreas Bulling.
\newblock Appearance-based gaze estimation in the wild.
\newblock In {\em 2015 IEEE Conference on Computer Vision and Pattern Recognition (CVPR)}, pages 4511--4520, 2015.

\bibitem{appearance_zhang_2}
Xucong Zhang, Yusuke Sugano, Mario Fritz, and Andreas Bulling.
\newblock Mpiigaze: Real-world dataset and deep appearance-based gaze estimation, 2017.

\bibitem{linear_regression_2}
Feng Lu, Takahiro Okabe, Yusuke Sugano, and Yoichi Sato.
\newblock A head pose-free approach for appearance-based gaze estimation.
\newblock In {\em British Machine Vision Conference}, 2011.

\bibitem{random_forest_1}
Yusuke Sugano, Yasuyuki Matsushita, and Yoichi Sato.
\newblock Learning-by-synthesis for appearance-based 3d gaze estimation.
\newblock In {\em 2014 IEEE Conference on Computer Vision and Pattern Recognition}, pages 1821--1828, 2014.

\bibitem{knn_1}
Erroll Wood, Tadas Baltru\v{s}aitis, Louis-Philippe Morency, Peter Robinson, and Andreas Bulling.
\newblock Learning an appearance-based gaze estimator from one million synthesised images.
\newblock In {\em Proceedings of the Ninth Biennial ACM Symposium on Eye Tracking Research \& Applications}, ETRA '16, page 131–138, 2016.

\bibitem{resnet_inception}
Rishi Athavale, Lakshmi~Sritan Motati, and Rohan Kalahasty.
\newblock One eye is all you need: Lightweight ensembles for gaze estimation with single encoders, 2022.

\bibitem{mazzeo2021openeds}
Pier~Luigi Mazzeo, Dilan D'Amico, Paolo Spagnolo, and Cosimo Distante.
\newblock Deep learning based eye gaze estimation and prediction.
\newblock In {\em 2021 6th International Conference on Smart and Sustainable Technologies (SpliTech)}, pages 1--6, 2021.

\bibitem{dosovitskiy2020image}
Alexey Dosovitskiy.
\newblock An image is worth 16x16 words: Transformers for image recognition at scale.
\newblock {\em arXiv preprint arXiv:2010.11929}, 2020.

\bibitem{nvgaze}
Joohwan Kim, Michael Stengel, Alexander Majercik, Shalini De~Mello, David Dunn, Samuli Laine, Morgan McGuire, and David Luebke.
\newblock Nvgaze: An anatomically-informed dataset for low-latency, near-eye gaze estimation.
\newblock In {\em Proceedings of the 2019 CHI Conference on Human Factors in Computing Systems}, CHI '19, page 1–12, New York, NY, USA, 2019. Association for Computing Machinery.

\bibitem{foveated_rendering_basic}
Anjul Patney, Marco Salvi, Joohwan Kim, Anton Kaplanyan, Chris Wyman, Nir Benty, David Luebke, and Aaron Lefohn.
\newblock Towards foveated rendering for gaze-tracked virtual reality.
\newblock {\em ACM Trans. Graph.}, 35(6), 2016.

\bibitem{latency_effect2}
Andrew~T. Duchowski, Donald~H. House, Jordan Gestring, Rui~I. Wang, Krzysztof Krejtz, Izabela Krejtz, Rados\l{}aw Mantiuk, and Bartosz Bazyluk.
\newblock Reducing visual discomfort of 3d stereoscopic displays with gaze-contingent depth-of-field.
\newblock In {\em Proceedings of the ACM Symposium on Applied Perception}, New York, NY, USA, 2014. Association for Computing Machinery.

\bibitem{latency_effect_3}
Rados Mantiuk, Bartosz Bazyluk, and Anna Tomaszewska.
\newblock Gaze-dependent depth-of-field effect rendering in virtual environments.
\newblock In {\em Proceedings of the Second International Conference on Serious Games Development and Applications}, Berlin, Heidelberg, 2011. Springer-Verlag.

\bibitem{ye2022rectanglefr}
Jiannan Ye, Anqi Xie, Susmija Jabbireddy, Yunchuan Li, Xubo Yang, and Xiaoxu Meng.
\newblock Rectangular mapping-based foveated rendering.
\newblock In {\em 2022 IEEE Conference on Virtual Reality and 3D User Interfaces (VR)}, pages 756--764, 2022.

\bibitem{foveal_definition}
M.~Weier, M.~Stengel, T.~Roth, P.~Didyk, E.~Eisemann, M.~Eisemann, S.~Grogorick, A.~Hinkenjann, E.~Kruijff, M.~Magnor, K.~Myszkowski, and P.~Slusallek.
\newblock Perception-driven accelerated rendering.
\newblock {\em Computer Graphics Forum}, 36(2):611--643, 2017.

\bibitem{nvidia_vrs}
NVIDIA Corporation.
\newblock Turing variable rate shading in vrworks.
\newblock \url{https://developer.nvidia.com/blog/turing-variable-rate-shading-vrworks/}, 2018.

\bibitem{VIT}
Alexey Dosovitskiy, Lucas Beyer, Alexander Kolesnikov, Dirk Weissenborn, Xiaohua Zhai, Thomas Unterthiner, Mostafa Dehghani, Matthias Minderer, Georg Heigold, Sylvain Gelly, Jakob Uszkoreit, and Neil Houlsby.
\newblock An image is worth 16x16 words: Transformers for image recognition at scale, 2021.

\bibitem{karpov2022vit}
Aleksei Karpov and Ilya Makarov.
\newblock Exploring efficiency of vision transformers for self-supervised monocular depth estimation.
\newblock In {\em 2022 IEEE International Symposium on Mixed and Augmented Reality (ISMAR)}, pages 711--719, 2022.

\bibitem{spvit}
Zhenglun Kong, Peiyan Dong, Xiaolong Ma, Xin Meng, Mengshu Sun, Wei Niu, Xuan Shen, Geng Yuan, Bin Ren, Minghai Qin, Hao Tang, and Yanzhi Wang.
\newblock Spvit: Enabling faster vision transformers via soft token pruning, 2022.

\bibitem{s2vite}
Tianlong Chen, Yu~Cheng, Zhe Gan, Lu~Yuan, Lei Zhang, and Zhangyang Wang.
\newblock Chasing sparsity in vision transformers: An end-to-end exploration, 2021.

\bibitem{evo-vit}
Yifan Xu, Zhijie Zhang, Mengdan Zhang, Kekai Sheng, Ke~Li, Weiming Dong, Liqing Zhang, Changsheng Xu, and Xing Sun.
\newblock Evo-vit: Slow-fast token evolution for dynamic vision transformer, 2021.

\bibitem{lancheres2019mipi}
Philippe Lancheres and Mohamed Hafed.
\newblock The mipi c-phy standard: A generalized multiconductor signaling scheme.
\newblock {\em IEEE Solid-State Circuits Magazine}, 11(2):69--77, 2019.

\bibitem{you2023eyecod}
Haoran You, Yang Zhao, Cheng Wan, Zhongzhi Yu, Yonggan Fu, Jiayi Yuan, Shang Wu, Shunyao Zhang, Yongan Zhang, Chaojian Li, et~al.
\newblock Eyecod: Eye tracking system acceleration via flatcam-based algorithm and hardware co-design.
\newblock {\em IEEE Micro}, 43(4):88--97, 2023.

\bibitem{liu20204}
Chiao Liu, Lyle Bainbridge, Andrew Berkovich, Song Chen, Wei Gao, Tsung-Hsun Tsai, Kazuya Mori, Rimon Ikeno, Masayuki Uno, Toshiyuki Isozaki, et~al.
\newblock A 4.6 $\mu$m, 512$\times$ 512, ultra-low power stacked digital pixel sensor with triple quantization and 127db dynamic range.
\newblock In {\em 2020 IEEE International Electron Devices Meeting (IEDM)}, pages 16--1. IEEE, 2020.

\bibitem{angelopoulos2020event}
Anastasios~N Angelopoulos, Julien~NP Martel, Amit~PS Kohli, Jorg Conradt, and Gordon Wetzstein.
\newblock Event based, near eye gaze tracking beyond 10,000 hz.
\newblock {\em arXiv preprint arXiv:2004.03577}, 2020.

\bibitem{sun2024estimating}
Xiaoyu Sun, Xiaochen Peng, Sai Zhang, Jorge Gomez, Win-San Khwa, Syed Sarwar, Ziyun Li, Weidong Cao, Zhao Wang, Chiao Liu, et~al.
\newblock Estimating power, performance, and area for on-sensor deployment of ar/vr workloads using an analytical framework.
\newblock {\em ACM Transactions on Design Automation of Electronic Systems}, 2024.

\bibitem{lee20196}
Pil-Ho Lee and Young-Chan Jang.
\newblock A 6.84 gbps/lane mipi c-phy transceiver bridge chip with level-dependent equalization.
\newblock {\em IEEE Transactions on Circuits and Systems II: Express Briefs}, 67(11):2672--2676, 2019.

\bibitem{mipi}
{What is Mobile Industry Processor Interface (MIPI) Protocol?}

\bibitem{power_quality_tradeoff}
Rahul Singh, Muhammad Huzaifa, Jeffrey Liu, Anjul Patney, Hashim Sharif, Yifan Zhao, and Sarita Adve.
\newblock Power, performance, and image quality tradeoffs in foveated rendering.
\newblock In {\em 2023 IEEE Conference Virtual Reality and 3D User Interfaces (VR)}, pages 205--214, 2023.

\bibitem{wang2004image}
Zhou Wang, Alan~C Bovik, Hamid~R Sheikh, and Eero~P Simoncelli.
\newblock Image quality assessment: from error visibility to structural similarity.
\newblock {\em IEEE transactions on image processing}, 13(4):600--612, 2004.

\bibitem{zhang2018perceptual}
Richard Zhang, Phillip Isola, Alexei~A Efros, Eli Shechtman, and Oliver Wang.
\newblock The unreasonable effectiveness of deep features as a perceptual metric.
\newblock In {\em CVPR}, 2018.

\bibitem{chen2024pea}
Kenneth Chen, Thomas Wan, Nathan Matsuda, Ajit Ninan, Alexandre Chapiro, and Qi~Sun.
\newblock Pea-pods: Perceptual evaluation of algorithms for power optimization in xr displays.
\newblock {\em ACM Transactions on Graphics (TOG)}, 43(4):1--17, 2024.

\bibitem{mantiuk2021fovvideovdp}
Rafa{\l}~K Mantiuk, Gyorgy Denes, Alexandre Chapiro, Anton Kaplanyan, Gizem Rufo, Romain Bachy, Trisha Lian, and Anjul Patney.
\newblock Fovvideovdp: A visible difference predictor for wide field-of-view video.
\newblock {\em ACM Transactions on Graphics (TOG)}, 40(4):1--19, 2021.

\bibitem{duinkharjav2022image}
Budmonde Duinkharjav, Praneeth Chakravarthula, Rachel Brown, Anjul Patney, and Qi~Sun.
\newblock Image features influence reaction time: A learned probabilistic perceptual model for saccade latency.
\newblock {\em ACM Transactions on Graphics (TOG)}, 41(4):1--15, 2022.

\bibitem{huang2023virtual}
Xincheng Huang, James Riddell, and Robert Xiao.
\newblock Virtual reality telepresence: 360-degree video streaming with edge-compute assisted static foveated compression.
\newblock {\em IEEE Transactions on Visualization and Computer Graphics}, 2023.

\bibitem{deng2022fov}
Nianchen Deng, Zhenyi He, Jiannan Ye, Budmonde Duinkharjav, Praneeth Chakravarthula, Xubo Yang, and Qi~Sun.
\newblock Fov-nerf: Foveated neural radiance fields for virtual reality.
\newblock {\em IEEE Transactions on Visualization and Computer Graphics}, 28(11):3854--3864, 2022.

\bibitem{bauer2022fovolnet}
David Bauer, Qi~Wu, and Kwan-Liu Ma.
\newblock Fovolnet: Fast volume rendering using foveated deep neural networks.
\newblock {\em IEEE transactions on visualization and computer graphics}, 29(1):515--525, 2022.

\bibitem{cui2022JND}
Dixuan Cui and Christos Mousas.
\newblock Estimating the just noticeable difference of tactile feedback in oculus quest 2 controllers.
\newblock In {\em 2022 IEEE International Symposium on Mixed and Augmented Reality (ISMAR)}, pages 1--7, 2022.

\bibitem{OpenEDS2020}
Cristina Palmero, Abhishek Sharma, Karsten Behrendt, Kapil Krishnakumar, Oleg~V. Komogortsev, and Sachin~S. Talathi.
\newblock Openeds2020: Open eyes dataset, 2020.

\bibitem{quest_pro}
Meta~Platform Inc.
\newblock Meta quest pro.
\newblock \url{https://www.meta.com/quest/quest-pro/}, 2022.

\bibitem{mobile_gpu}
Quadro rtx.
\newblock \url{https://www.nvidia.com/en-us/design-visualization/rtx/}, 2022.

\bibitem{DCS}
{Eagle Dynamics}.
\newblock {Digital Combat Simulator}.
\newblock \url{https://www.digitalcombatsimulator.com/en/}, 2008.

\bibitem{zielinski2015exploring}
David~J Zielinski, Hrishikesh~M Rao, Mark~A Sommer, and Regis Kopper.
\newblock Exploring the effects of image persistence in low frame rate virtual environments.
\newblock In {\em 2015 IEEE Virtual Reality (VR)}, pages 19--26. IEEE, 2015.

\bibitem{apple_vision_pro}
{Apple Inc.}
\newblock {Apple Vision Pro}, 2024.

\bibitem{pupil_2013}
W.~Sprague, Zachary Helft, Jared Parnell, J.~Schmoll, G.~Love, and Martin Banks.
\newblock Pupil shape is adaptive for many species.
\newblock {\em Journal of Vision}, 13:607--607, 07 2013.

\bibitem{wikipedia_roundness}
{Wikipedia contributors}.
\newblock Roundness.
\newblock \url{https://en.wikipedia.org/wiki/Roundness}, 2024.

\bibitem{code_github}
Horizon Research and PyDSGZ.
\newblock Edgaze: Efficient gaze tracking and deepvog: Deep learning for eye tracking in vr.
\newblock \url{https://github.com/horizon-research/edgaze} and \url{https://github.com/pydsgz/DeepVOG}, 2022.
\newblock Accessed: 2024-09-14.

\bibitem{lin2014microsoft}
Tsung-Yi Lin, Michael Maire, Serge Belongie, James Hays, Pietro Perona, Deva Ramanan, Piotr Doll{\'a}r, and C~Lawrence Zitnick.
\newblock Microsoft coco: Common objects in context.
\newblock In {\em Computer Vision--ECCV 2014: 13th European Conference, Zurich, Switzerland, September 6-12, 2014, Proceedings, Part V 13}, pages 740--755. Springer, 2014.

\bibitem{perry2002gaze}
Jeffrey~S Perry and Wilson~S Geisler.
\newblock Gaze-contingent real-time simulation of arbitrary visual fields.
\newblock In {\em Human vision and electronic imaging VII}, volume 4662, pages 57--69. SPIE, 2002.

\bibitem{jiang2015salicon}
Ming Jiang, Shengsheng Huang, Juanyong Duan, and Qi~Zhao.
\newblock Salicon: Saliency in context.
\newblock In {\em Proceedings of the IEEE conference on computer vision and pattern recognition}, pages 1072--1080, 2015.

\bibitem{gpusim}
Gpgpu-sim.
\newblock \url{https://github.com/gpgpu-sim/gpgpu-sim_distribution}.

\bibitem{adreno}
Adreno gpu.
\newblock \url{https://www.notebookcheck.net/Qualcomm-Adreno-650-GPU-Benchmarks-and-Specs.448196.0.html}.

\end{thebibliography}

\end{document}